\documentclass[aos,noinfoline]{imsart}
\usepackage{fullpage}
\usepackage{times}
\usepackage{amsmath,amssymb}
\usepackage{graphicx}
\usepackage{url}
\usepackage{xfrac}
\usepackage{booktabs,siunitx}
\usepackage[caption=false]{subfig}
\usepackage[labelformat=simple,justification=raggedright,format=hang,font={small,it},singlelinecheck=false]{caption}
\usepackage{macros}
\usepackage{floatrow}
\usepackage{titlesec}
\usepackage{algorithm}
\usepackage{algorithmic}
\usepackage{natbib}

\setattribute{journal}{name}{}

\titleformat{\section}[hang]
  {\bfseries}{\thesection.}{1em}{}

\begin{document}
\runtitle{Freeze-Thaw Bayesian Optimization}
\title{Freeze-Thaw Bayesian Optimization}

\begin{aug}
  \author{\fnms{Kevin} \snm{Swersky}\ead[label=e2]{kswersky@cs.toronto.edu}},
  \author{\fnms{Jasper}  \snm{Snoek}\ead[label=e1]{jsnoek@seas.harvard.edu}}
  \and
  \author{\fnms{Ryan P.}  \snm{Adams}\ead[label=e4]{rpa@seas.harvard.edu}}

  \runauthor{Swersky et al.}

  \affiliation{Harvard University and University of Toronto}

  \address{Kevin Swersky\\
   		Department of Computer Science\\
        University of Toronto\\
        \printead{e2}}

  \address{Jasper Snoek\\
   		School of Engineering and Applied Sciences\\
        Harvard University\\
        \printead{e1}}

\address{Ryan P. Adams\\
   		School of Engineering and Applied Sciences\\
        Harvard University\\
        \printead{e4}}
  
\end{aug}

\begin{abstract}
In this paper we develop a dynamic form of Bayesian optimization for machine learning models with the goal of rapidly finding good hyperparameter settings. Our method uses the partial information gained during the training of a machine learning model in order to decide whether to pause training and start a new model, or resume the training of a previously-considered model. We specifically tailor our method to machine learning problems by developing a novel positive-definite covariance kernel to capture a variety of training curves. Furthermore, we develop a Gaussian process prior that scales gracefully with additional temporal observations. Finally, we provide an information-theoretic framework to automate the decision process. Experiments on several common machine learning models show that our approach is extremely effective in practice.
\end{abstract}

\section{Introduction}
In machine learning, the term ``training'' is used to describe the procedure of fitting a model to data.  In many popular models, this fitting procedure is framed as an optimization problem, in which a loss is minimized as a function of the parameters.  In all but the simplest machine learning models, this minimization must be performed with an iterative algorithm such as stochastic gradient descent or the nonlinear conjugate gradient method.

Another aspect of training involves fitting model ``hyperparameters.'' These are parameters that in some way govern the model space or fitting procedure; in both cases they are typically difficult to minimize directly in terms of the training loss and are usually evaluated in terms of generalization performance via held-out data.  Hyperparameters are often regularization penalties such as~$\ell_p$ norms on model parameters, but can also capture model capacity as in the number of hidden units in a neural network.  These hyperparameters help determine the appropriate bias-variance tradeoff for a given model family and data set.  On the other hand, hyperparameters of the fitting procedure govern algorithmic aspects of training, such as the learning rate schedule of stochastic gradient descent, or the width of a Monte Carlo proposal distribution.  The goal of fitting both kinds of hyperparameters is to identify a model and an optimization procedure in which successful minimization of training loss is likely to result in good generalization performance.  When a held-out validation set is used to evaluate the quality of hyperparameters, the overall optimization proceeds as a double loop, where the outer loop sets the hyperparameters and the inner loop applies an iterative training procedure to fit the model to data.

Often this outer hyperparameter optimization is performed by hand, which---even if rigorously systematized---can be a difficult and laborious process.  Simple alternatives include the application of heuristics and intuition, grid search, which scales poorly with dimension, or random search~\cite{bergstra-bengio-2012a}, which is computationally expensive due to the need to train many models.  In light of this, Bayesian optimization~\cite{Mockus1978} has recently been proposed as an effective method for systematically and intelligently setting the hyperparameters of machine learning models~\cite{BergstraJ2011,snoek-etal-2012b}. Using a principled characterization of model uncertainty, Bayesian optimization attempts to find the best hyperparameter settings with as few model evaluations as possible.

One issue with previously proposed approaches to Bayesian optimization for machine learning is that a model must be fully trained before the quality of its hyperparameters can be assessed.  Human experts, however, appear to be able to rapidly assess whether or not a model is likely to \emph{eventually} be useful, even when the inner-loop training is only partially completed.  When such an assessment can be made accurately, it is possible to explore the hyperparameter space more effectively by aborting model fits that are likely to be low quality.  The goal of this paper is to take advantage of the partial information provided by iterative training procedures, within the Bayesian optimization framework for hyperparameter search.  We propose a new technique that makes it possible to estimate when to pause the training of one model in favor of starting a new one with different hyperparameters, or resuming a partially-completed training procedure from an old model.  We refer to our approach as \emph{freeze-thaw} Bayesian optimization, as the algorithm maintains a set of ``frozen'' (partially completed but not being actively trained) models and uses an information-theoretic criterion to determine which ones to ``thaw'' and continue training.

Our approach hinges on the assumption that, for many models, the training loss during the fitting procedure roughly follows an exponential decay towards an unknown final value.   We build a Bayesian nonparametric prior around this assumption by developing a new kernel that is an infinite mixture of exponentially-decaying basis functions, with the goal of characterizing these training curves. Using this kernel with a novel and efficient temporal Gaussian process prior, we are able to forecast the final result of partially trained models and use this during Bayesian optimization to determine the most promising action. We demonstrate that freeze-thaw Bayesian optimization can find good hyperparameter settings for many different models in considerably less time than ordinary Bayesian optimization.

\section{Background}
\subsection{Gaussian Processes}
A Gaussian process (GP) is a probability distribution over the space of functions ${f: \mathcal{X} \mapsto \mathbb{R}}$. It is fully specified by a kernel function ${k(\brmx,\brmx'): \mathcal{X}\times\mathcal{X} \mapsto \mathbb{R}}$ and a mean function ${m(\brmx): \mathcal{X} \mapsto \mathbb{R}}$. The input space~$\mathcal{X}$ can be anything for which there exists a positive definite kernel.  Gaussian processes are often used as prior distributions over functions. Given input/output pairs $ \{ (\brmx_n,y_n) \}_{n=1}^N$, we can express the GP posterior predictive distribution at a new input $\brmx_*$ by
\begin{align}
P(f_*; \{ (\brmx,\brmy) \}_{n=1}^N, \brmx_*) &= \mathcal{N}(f_*;\mu_*,v_*), \nonumber \\
\mu_* &= m(\brmx_*) + \brmK_* \brmK^{-1} (\brmy - \brmm), \\
v_* &= \brmK_{**} - \brmK_* \brmK^{-1} \brmK_*^\top\,.
\end{align}
Here $\brmK$ is the Gram matrix formed by applying the kernel function to every point in the dataset,~$\brmK_*$  is a vector formed by computing $k(\brmx_*,\brmx_n)$ for all $n$ in the dataset, ${\brmK_{**} = k(\brmx_*,\brmx_*)}$ and $\brmm$ is the vector formed by applying the mean function to all inputs in the dataset.

GPs have become a mainstay of Bayesian nonparametric regression because of their simplicity, flexibility, and ability to characterize uncertainty when making predictions. This comes at a computational cost that grows as $O(N^3)$ due to the requirement of inverting the Gram matrix.

Implicit in the mean and kernel functions are GP hyperparameters $\theta$ (not to be confused with the hyperparameters we are optimizing) that characterize the overall behaviour of the GP. We follow the same procedure as in~\cite{snoek-etal-2012b}, using a Mat\'{e}rn kernel, an inferred constant prior mean, and integrating out the hyperparameters via slice sampling~\cite{Murray-Adams-2010a}.

\subsection{Bayesian Optimization}
Bayesian optimization is a methodology for the global optimization of expensive, noisy functions over a bounded domain; without loss of generality, we consider the domain~$\mathcal{X}$ to be the unit hypercube~${[0,1]^D}$. Bayesian optimization has received significant interest recently in machine learning due in part to the maturing of methods for Bayesian nonparametric regression.  Recent work has developed compelling theoretical convergence guarantees~\cite{Srinivas2010,Bull2011,defreitas-etal-2013a}, more rigorous methodologies~\cite{osborne-2009a,hoffman-etal-2011} and methods tailored to specific applications such as hyperparameter optimization~\cite{hutter-2011a,snoek-etal-2012b,BergstraJ2011} and sensor set selection~\cite{garnett-etal-2010}.  Bayesian optimization (see \cite{brochu-etal-2010a,lizotte-thesis} for an in-depth explanation) proceeds by fitting a probabilistic model to the data and then using this model as an inexpensive proxy in order to determine the most promising location to evaluate next.

We assume that the goal is to minimize the objective.  The choice of which point, $\brmx_{*}$, to evaluate next is determined by maximizing an acquisition function,~${\brmx_* = \argmax_{\brmx}a(\brmx)}$. This is a utility function that determines the trade-off between exploring in regions the model is uncertain about, and exploiting in regions that are likely to yield a good result.

A popular and effective acquisition function is the expected improvement (EI) criterion~\cite{Jones2001,Mockus1994},
\begin{align}
a_{\mathrm{EI}}(\brmx) &= \sqrt{v(\brmx)}(\gamma(\brmx)\Phi(\gamma(\brmx)) + \phi(\gamma(\brmx)), &
\gamma(\brmx) = \frac{f(\brmx_{\mathrm{best}}-\mu(\brmx)))}{\sqrt{v(\brmx)}}, \label{eq:ei}
\end{align}
where $\phi$ and $\Phi$, respectively, are the probability density function and cumulative distribution function of the standard normal distribution, $\brmx_{\mathrm{best}}$ is the input corresponding to the minimum output observed so far and $\mu(\brmx)$ and $v(\brmx)$ are the posterior mean and variance of the probabilistic model evaluated at $\brmx$.

While EI focuses on the value of the function, an alternative approach is to consider information gathered about the location of the minimum.  Given a posterior predictive distribution over the minimum, the \emph{entropy search} \cite{hennig-schuler-2012} approach at each step chooses the input that most reduces expected uncertainty over the location of the minimum.  Let $P_{\mathrm{min}}$ represent the current estimated distribution over the minimum. The entropy search acquisition function is given by
\begin{align}
a_{\mathrm{ES}}(\brmx) &= \int (H(P_{\mathrm{min}}^{y}) - H(P_{\mathrm{min}})) P(y \given \{(\brmx_n,y_n) \}_{n=1}^N)\,dy.
\end{align}
Here $P_{\mathrm{min}}^{y}$ is an updated distribution over the location of the minimum given that point $\mathrm{\brmx}$ yields the observation $y$. In other words, entropy search maximizes the expected information gain over the location of the minimum from evaluating a point. In practice, we follow~\cite{swersky-etal-2013a,gelbart-etal-2014} and use Monte Carlo simulation to estimate $P_{\mathrm{min}}$ on a discrete grid containing the points with the highest EI.

\begin{figure}[t]
\centering%
\subfloat[\label{fig:decay_basis}Exponential Decay Basis]{%
  \includegraphics[width=0.32\textwidth]{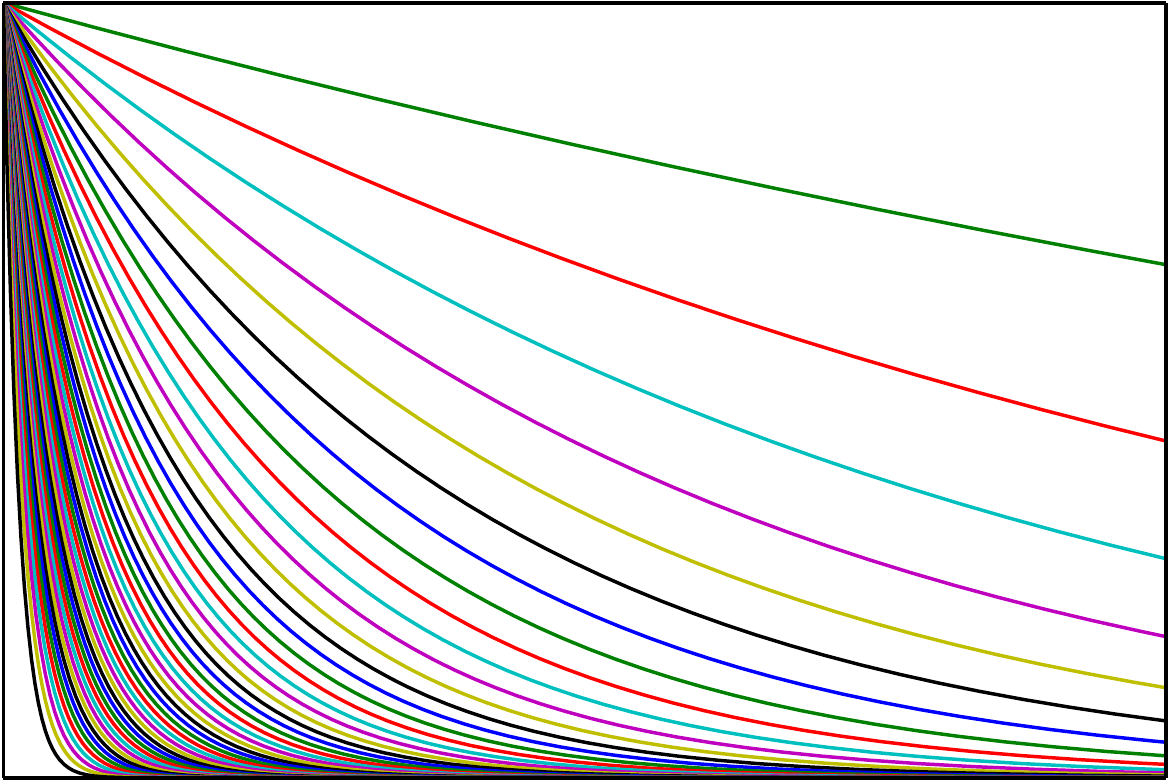}}\hfill
\subfloat[\label{fig:decay_samples}Samples]{%
  \includegraphics[width=0.32\textwidth]{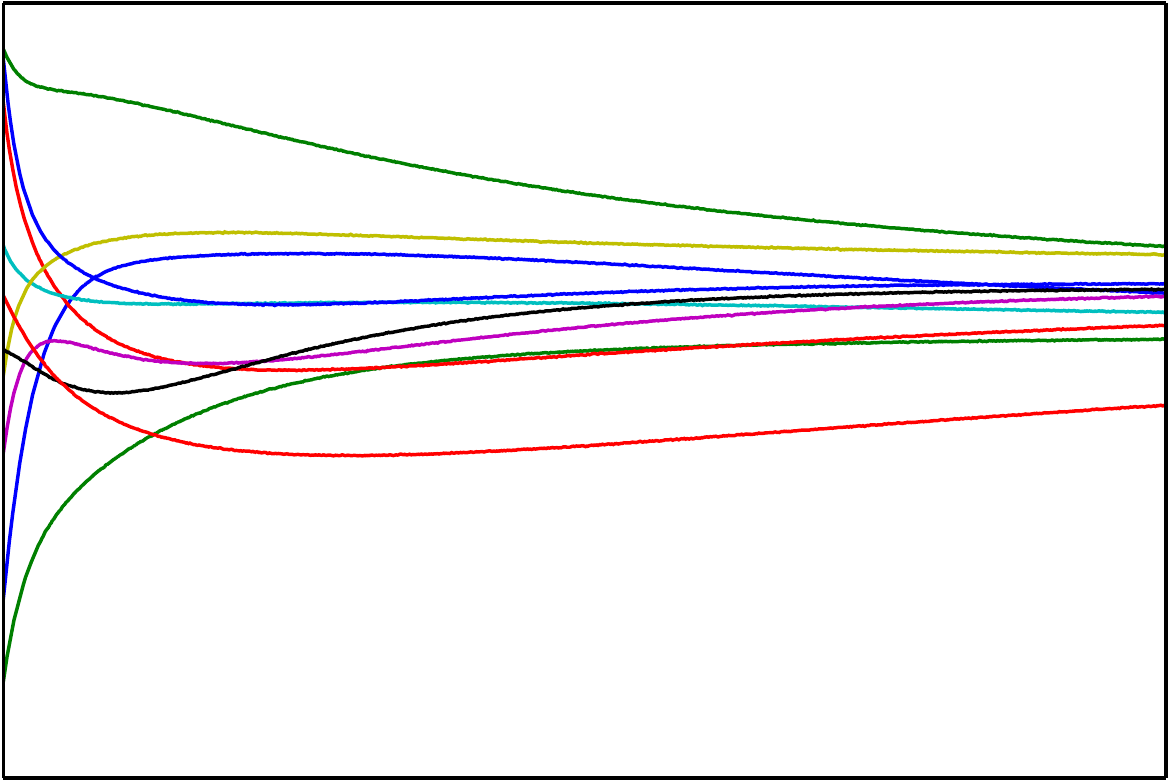}}\hfill
\subfloat[\label{fig:decay_curves}Training Curve Samples]{%
  \includegraphics[width=0.32\textwidth]{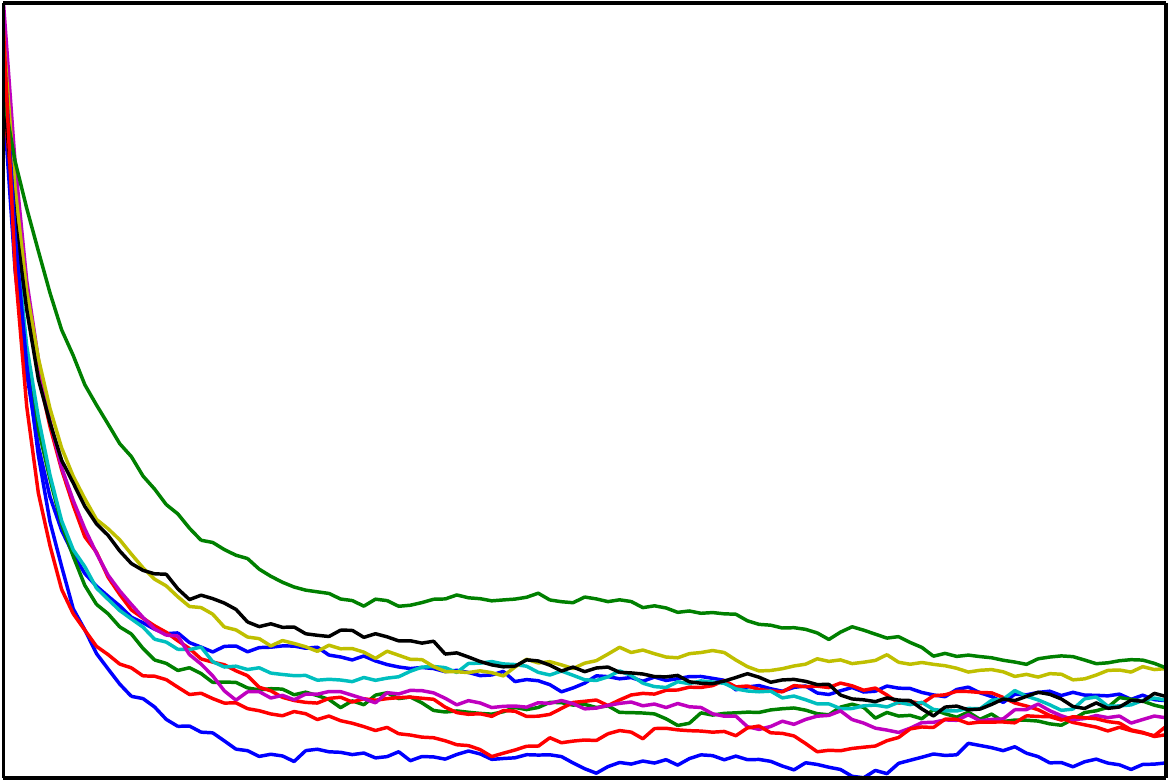}}
\caption{\label{fig:kernel}Example functions from the exponential decay kernel. (a)~Example functions from our basis set with $\alpha=1.0$ and $\beta=0.5$.  (b)~Samples from a Gaussian process with this covariance function. (c)~Samples from a Gaussian process conditioned on the curves starting at a positive number and with an added Ornstein-Uhlenbeck kernel to simulate natural training curves.}
\end{figure}

\section{A Kernel for Training Curves}
\label{sec:kernel}
We develop here a positive definite covariance kernel designed to model iterative optimization curves.  Specifically, we develop a prior that strongly supports exponentially decaying functions of the form~$e^{-\lambda t}$ for~${t,\lambda\geq 0}$. Rather than assume a finite basis with a fixed set of~$\lambda$ terms, we integrate over infinite basis functions parameterized by $\lambda$ from $0$ to $\infty$ with a mixing measure that allows us to weight regions of the range.  Thus the covariance function $k(t,t')$ between two inputs $t$ and $t'$ is given by:
\begin{equation}
\label{eqn:expcov}
k(t,t') = \int_{0}^{\infty} e^{-\lambda t} e^{-\lambda t'} \psi(d\lambda)\,,
\end{equation}
where~$\psi$ is a non-negative mixing measure on~$\reals_+$.  It is particularly convenient to take~$\psi$ to have the form of a gamma distribution with density~${\psi(\lambda)=\frac{\beta^{\alpha}}{\Gamma(\alpha)}\lambda^{\alpha-1}e^{-\lambda \beta}}$, for parameters~${\alpha,\beta>0}$ and in which $\Gamma(\cdot)$ is the gamma function.  This choice of mixing measure leads to an analytic solution for Equation \ref{eqn:expcov}:
\begin{align}
k(t,t') & = \int_{0}^{\infty} e^{-\lambda (t + t')} \frac{\beta^{\alpha}}{\Gamma(\alpha)} \lambda^{\alpha-1}e^{-\lambda \beta}\,d\lambda \notag\\
&= \frac{\beta^{\alpha}}{\Gamma(\alpha)} \int_{0}^{\infty} e^{-\lambda (t + t' + \beta)} \lambda^{\alpha-1}\, d\lambda
= \frac{\beta^{\alpha}}{(t + t' + \beta)^{\alpha}}.
\end{align}

In Figure~\ref{fig:kernel}, we show visualizations of the basis set, samples from a Gaussian process prior with this covariance function and samples from a model specifically for optimization curves.  In the following, we use this kernel in our model as the covariance function over time steps for an iterative optimization procedure being modeled. For noisy curves, this kernel can be composed with e.g., a noise kernel or an Ornstein-Uhlenbeck kernel.

\section{Efficient Gaussian Processes for Iterative Training Procedures}
\subsection{Specification}
\begin{figure}[t]
	\centering
	\subfloat[\label{fig:spatiotemporalgp}Graphical Model]{%
		\includegraphics[width=0.33\textwidth]{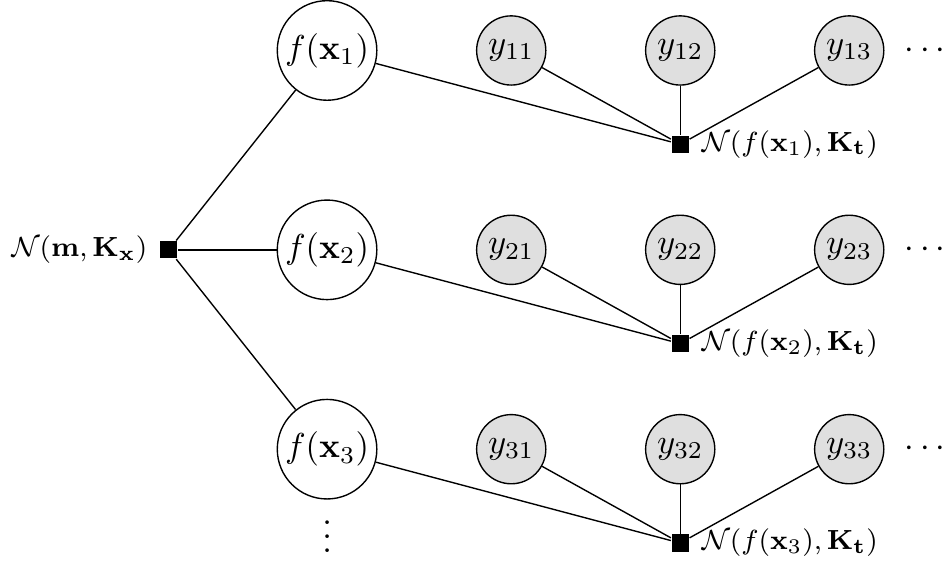}
	}
	\subfloat[\label{fig:temporalpreds}Training curve predictions]{%
		\includegraphics[width=0.28\textwidth]{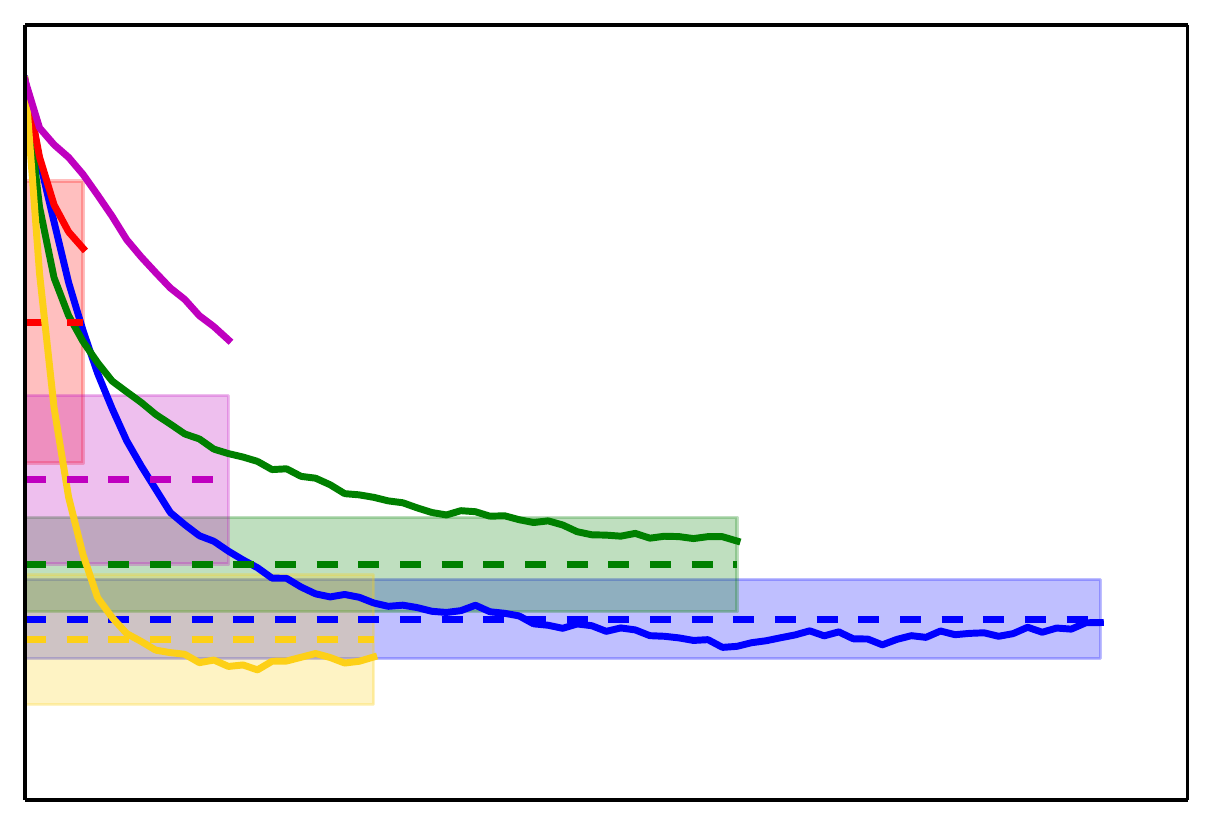}
	}
	\subfloat[\label{fig:spatialpreds}Asymptotic GP]{%
		\includegraphics[width=0.28\textwidth]{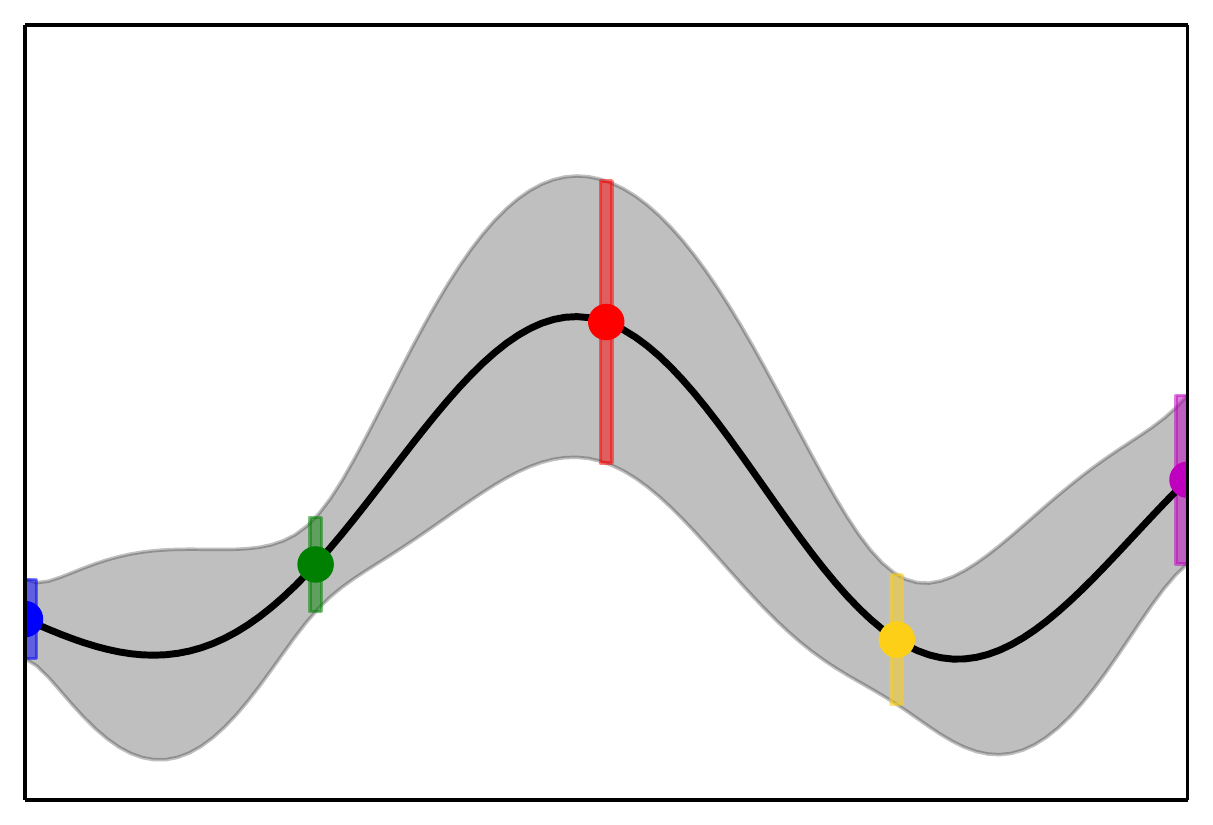}
	}
	\caption{\protect\subref{fig:spatiotemporalgp} Factor graph representation of the GP model for training procedures. Each row represents a learning curve that is drawn from an independent GP prior, conditioned on its mean. The mean of each learning curve is jointly drawn with the mean of the other curves using another GP prior. \protect\subref{fig:temporalpreds}  Partially completed training curves, and the GP prediction at their eventual asymptotes.  \protect\subref{fig:spatialpreds} The posterior GP prediction at the asymptote. Each colored point represents the GP prediction at the hyperparameter location corresponding to a training curve with the same color.}
\end{figure}
In order to perform Bayesian optimization we need to use a surrogate model as a proxy for the function of interest. The main issue with GPs is that making predictions requires $O(N^3)$ computation, where $N$ is the number of training observations. In the context of freeze-thaw Bayesian optimization, a na\"{i}ve model would put a Gaussian process prior over every observed training loss through time. With $N$ unique hyperparameters settings and up to $T$ training iterations per setting, computing quantities of interest using a GP would scale as $O(N^3T^3)$, which is prohibitively expensive.

To reduce the computational cost, we incorporate a conditional independence assumption that each training curve is drawn from a separate Gaussian process, conditioned on the prior mean, which is itself drawn from a global Gaussian process.  We define~$\brmy_n$ to be a vector of generalization loss measurements for the~$n$th hyperparameter setting, i.e., it is a time series.  Formally, we model the distribution over these 
training curves $\{\brmy_n \}_{n=1}^N$ given hyperparameter settings $\{\brmx_n\}_{n=1}^N$ as,
\begin{align}
P(\{\brmy_n \}_{n=1}^N \given \{\brmx_n\}_{n=1}^N) &= \int \left [ \prod_{n=1}^N \distNorm(\brmy_n; f_n\ovec_n, \brmK_{\mathrm{t}n}) \right ] \distNorm(\brmf; \brmm, \brmK_\rmx)\, d\brmf,
\end{align}
where $\brmf$ is a latent function that specifies the mean of each training curve and $\ovec$ is a column vector of~$1$'s. In other words, the generative procedure is to first draw a latent function $\brmf$ over hyper parameter settings according to a GP prior. Conditioned on $\brmf$, each training curve is modelled independently using another GP prior with a mean given by the corresponding entry of $\brmf$. We use a constant mean $\brmm$ (which we infer) and Mat\'{e}rn-5/2 kernel for the GP prior over hyperparameters. Each time-based GP uses the covariance given in Section \ref{sec:kernel}. A graphical illustration of this model is shown in Figure \ref{fig:spatiotemporalgp}.

The training curves will asymptotically converge to $\brmf$ away from the observed points. As we will demonstrate, the conditional independence assumption is not too restrictive for the kinds of training curves that are typically found when training machine learning models.

Using properties of the Gaussian distribution, we can write the joint distribution over $\brmy$ and $\brmf$ as,
\begin{align}
P(\brmy,\brmf \,|\, \{ \brmx_n \}_{n=1}^N) &= \distNorm \left ( 
	\left ( \begin{array}{c}
	 	\brmf \\
		\brmy
	\end{array} \right )
	; \left ( \begin{array}{c}
	 	\mathbf{\brmm} \\
		\mathbf{O\brmm}
	\end{array} \right )
	\left ( \begin{array}{cc}
		\brmK_\rmx & \brmK_\rmx \brmO^\top \\
		\brmO\brmK_\rmx & \brmK_\rmt + \brmO\brmK_\rmx \brmO^\top
	\end{array} \right )
\right ),
\label{eq:gpjointlik}
\end{align}
where $\brmy=(\brmy_1, \brmy_2, \ldots, \brmy_N)^\top$, $\brmO = \mathrm{blockdiag}(\ovec_1,\ovec_2, \ldots, \ovec_N)$ is a block-diagonal matrix, where each block is a vector of ones corresponding to number of observations in its corresponding training curve, and $\brmK_\rmt = \mathrm{blockdiag}(\brmK_{\rmt 1}, \brmK_{\rmt 2}, \ldots, \brmK_{\rmt N})$ is a block-diagonal matrix where each block is the covariance for it's corresponding training curve.

\subsection{Inference}
Using this representation, we can efficiently compute the required quantities for Bayesian optimization. We have omitted some of the details of these derivations; they can be found in the appendix.

\paragraph{Marginal likelihood}
The marginal likelihood is required to estimate the hyperparameters of the GP. Using the marginalization property of the Gaussian distribution, the marginal likelihood can be derived from Equation \ref{eq:gpjointlik} and is given by,
\begin{align}
	P(\brmy \,|\, \{ \brmx_n \}_{n=1}^N) &= \distNorm \left ( \brmy ; \brmO \brmm, \brmK_\rmt + \brmO\brmK_\rmx \brmO^\top \right ).
\end{align}
The covariance of this distribution has a size of $NT\times NT$, however we can efficiently invert it using the Woodbury matrix identity.
\begin{align}
(\brmK_\rmt + \brmO\brmK_\rmx \brmO^\top)^{-1} &= \brmK_\rmt^{-1} - \brmK_\rmt^{-1}\brmO(\brmK_\rmx^{-1} + \brmO^\top\brmK_\rmt^{-1}\brmO)^{-1}\brmO^{\top}\brmK_\rmt^{-1}.
\end{align}
We can also use the analogous matrix determinant lemma to obtain an efficient representation for the normalization constant of the Gaussian distribution, allowing us to write the log-likelihood as,
\begin{align}
\log P(\brmy \,|\, \{ \brmx_n \}_{n=1}^N) &= -\frac{1}{2} (\brmy - \brmO \brmm)^\top \brmK_{\rmt}^{-1} (\brmy - \brmO \brmm) + \frac{1}{2} \bgamma^\top (\brmK_\rmx^{-1} + \bLambda)^{-1} \bgamma \nonumber \\
&\quad-\frac{1}{2}\left ( \log \left (|\brmK_\rmx^{-1} + \bLambda |\right ) + \log \left (|\brmK_\rmx|\right ) + \log \left (|\brmK_\rmt| \right )\right ) + const
\end{align}
Where $\bgamma = \brmO^\top \brmK_\rmt^{-1}(\brmy - \brmO \brmm)$ and a specific element can be written as, $\bgamma_n = \ovec_n^\top \brmK_{\rmt n}^{-1} (\brmy_n - \brmm_n)$; $\bLambda=\brmO^\top \brmK_\rmt^{-1} \brmO = \mathrm{diag}(\lambda_1,\lambda_2,\ldots,\lambda_N)$, where $\lambda_n=\ovec_n^\top \brmK_{\rmt n}^{-1} \ovec_n$.

\paragraph{Posterior distribution}
Using the conditioning property of the Gaussian, we can express the posterior $P(\brmf \,|\, \brmy, \{ \brmx_n \}_{n=1}^N)$ as,
\begin{align}
P(\brmf \,|\, \brmy, \{ \brmx_n \}_{n=1}^N) &= \distNorm \left ( \brmf ; \bmu, \brmC \right ), \nonumber \\
\bmu &= \brmm + \brmK_\rmx \bgamma - \brmK_\rmx \bLambda (\brmK_\rmx^{-1} + \bLambda)^{-1} \bgamma, \nonumber \\
&= \brmm + \brmC \bgamma, \\
\brmC &= \brmK_\rmx - \brmK_\rmx \bLambda \brmK_\rmx + \brmK_\rmx \bLambda (\brmK_\rmx^{-1} + \bLambda)^{-1} \bLambda \brmK_\rmx, \nonumber \\
&= \brmK_\rmx - \brmK_\rmx(\brmK_\rmx + \bLambda^{-1})^{-1} \brmK_\rmx.
\end{align}

\paragraph{Posterior predictive distributions}
Given a new hyperparameter setting $\brmx_*$, the posterior distribution $P(f_*\,|\,\brmy, \{ \brmx_n \}_{n=1}^N, \brmx_*)$ is given by,
\begin{align}
P(f_* \,|\, \brmy, \{ \brmx_n \}_{n=1}^N, \brmx_*) &= \int P(f_* \,|\, \brmf, \brmx_*)P(\brmf \,|\, \brmy, \{ \brmx_n \}_{n=1}^N) d\brmf, \nonumber \\
&= \int \distNorm (f_* ; m + \brmK_{\rmx *}^\top \brmK_\rmx^{-1} \brmf, \brmK_{\rmx **} - \brmK_{\rmx *}^\top \brmK_\rmx^{-1} \brmK_{\rmx *}) \distNorm (\brmf ; \bmu, \brmC) d\brmf, \nonumber \\
&= \distNorm (f_* ; m\! +\! \brmK_{\rmx *}^\top \brmK_\rmx^{-1} (\bmu\! -\! \brmm), \brmK_{\rmx **}\! -\! \brmK_{\rmx *}^\top (\brmK_\rmx\! +\! \bLambda^{-1})^{-1}\brmK_{\rmx *}). \label{eq:postpred}
\end{align}
Similarly, the posterior predictive distribution for a new point in a training curve, $y_{n*}$,  is given by,
\begin{align}
P(y_{n*}\,|\,  \{ \brmx_n \}_{n=1}^N, \brmy) &= \int P(y_{n*} \,|\, \brmy_n, \brmf_n) P(\brmf_n \,|\, \brmy, \{ \brmx_n \}_{n=1}^N) d\brmf_n, \nonumber \\
&= \int \distNorm(y_{n*} ; \brmf_n \ovec_* + \brmK_{\rmt n *}^\top \brmK_{\rmt n}^{-1} (\brmy_n - \ovec_n\brmf_n), \brmK_{\rmt n **} - \brmK_{\rmt n *}^\top \brmK_{\rmt n}^{-1} \brmK_{\rmt n *}) \nonumber \\ &\qquad\ \distNorm(\brmf_n; \bmu_n, \brmC_{nn}) d\brmf_n, \nonumber \\
&= \distNorm(y_{n*} ; \brmK_{\rmt n *}^\top \brmK_{\rmt n}^{-1} \brmy_n + \Omega \bmu_n, \brmK_{\rmt n **} - \brmK_{\rmt n *}^\top \brmK_{\rmt n}^{-1} \brmK_{\rmt n *} + \Omega \brmC_{nn} \Omega^\top), \label{eq:obsvpred} \\
\Omega &= \ovec_* - \brmK_{\rmt n *}^\top \brmK_{\rmt n}^{-1} \ovec_n. \nonumber
\end{align}
Where $\ovec_*$ is a vector of ones with size equal to the number of time-steps we are predicting. We have omitted the dependence on $t$ for brevity since it is a regularly spaced grid in our experiments.

Finally, in the absence of any observations, the posterior predictive distribution is given by,
\begin{align}
P(y_*\,|\,  \{ \brmx_n \}_{n=1}^N, \brmy, \brmx_*) &=  \distNorm(y_{n*} ; \bmu_*, \brmK_{\rmt **} + \Sigma_{**}),
\label{eq:noobsvpred}
\end{align}
Where $\mu_*$ and $\Sigma_{**}$ are the mean and variance given by Equation \ref{eq:postpred}.
\subsection{Computational Complexity} In order to evaluate the GP we need to invert $\brmK_\rmx$ and $\brmK_\rmt$ independently. When computing the quantities $\bLambda$ and $\bgamma$ we can pre-compute the Cholesky decomposition of $\brmK_\rmt$ and use this to solve linear systems with a $T \times T$ matrix $N$ times, leading to a total computational complexity of $O(N^3 + T^3 + NT^2)$. In practice $T$ is somewhere between $10$ and $100$, or can be kept small by sampling the training curves in coarser increments.

\begin{algorithm}[t!]
\caption{Entropy Search Freeze-Thaw Bayesian Optimization}
\label{alg:entropysearch}
\begin{algorithmic}[1]
\STATE Given a basket $\{(\brmx,\brmy) \}_{B_\mathrm{old}} \cup \{(\brmx)\}_{B_\mathrm{new}}$
\STATE $a = (0,0,\ldots,0)$
\STATE Compute $P_\mathrm{min}$ over the basket using Monte Carlo simulation and Equation \ref{eq:postpred}.
\FOR{each point $\brmx_k$ in the basket}
\STATE // $n_\mathrm{fant}$ is some specified number, e.g., $5$.
\FOR{$i=1\ldots n_{\mathrm{fant}}$}
\IF{the point is old}
\STATE Fantasize an observation $y_{t+1}$ using Equation \ref{eq:obsvpred}.
\ENDIF
\IF{the point is new}
\STATE Fantasize an observation $y_1$ using Equation \ref{eq:noobsvpred}.
\ENDIF
\STATE Conditioned on this observation, compute $P_\mathrm{min}^y$ over the basket using Monte Carlo simulation and Equation \ref{eq:postpred}.
\STATE $a(k) \leftarrow a(k) + \frac{H(P_\mathrm{min}^y) - H(P_\mathrm{min})}{n_\mathrm{fant}}$ // information gain.
\ENDFOR
\ENDFOR
\STATE Select $\brmx_k$, where $k = \argmax_k a(k)$ as the next model to run.
\end{algorithmic}
\end{algorithm}

\section{Bayesian Optimization for Iterative Training Procedures}
Using the GP developed in the previous sections, our goal is to create an automatic system that can intelligently decide when to pause training of current models, resume training of previous models, or start new models for training. The optimal strategy is critically dependent on the goal of the user, and we assume that this is to find the best model. That is, if every model were to be fully trained, then the one with the lowest asymptotic error is the one we want to discover. This is reflected in our GP, which becomes a standard GP over hyperparameter settings at the asymptote of each training curve.

Our Bayesian optimization strategy proceeds by maintaining a basket of $B=B_{\mathrm{old}} + B_{\mathrm{new}}$ candidate models. $B_{\mathrm{old}}$ represents some number of models that have already been trained to some degree, while $B_{\mathrm{new}}$ represents some number of brand new models. In practice, we set $B_{\mathrm{old}}=10$ and $B_{\mathrm{new}}=3$. The entire basket is chosen using models with the maximum EI at the asymptote, which is computed using Equations \ref{eq:postpred} and \ref{eq:ei}. Each round, after a new observation has been collected, the basket is re-built using possibly different models. This step is essentially standard Bayesian optimization using EI.

Given a basket, the task now becomes one of choosing which candidate to run. Naively choosing EI would always favor picking new models rather than running old ones for more iterations since the new models have maximal EI by definition. Instead, similar to~\cite{swersky-etal-2013a}, we use the entropy search acquisition function to pick the point that provides the most information about the location of the minimum at the asymptote. The method is summarized in Algorithm \ref{alg:entropysearch}. This procedure is similar to standard entropy search, except that here we are not just fantasizing outcomes for unseen inputs but also fantasizing outcomes for already seen inputs. Since we are considering the results at the asymptote, each subsequent observation provides more information about the true function.

\begin{figure}[t!]
\centering%
\subfloat[\label{fig:logistic}Logistic Regression]{%
  \includegraphics[width=0.32\textwidth,clip=true,trim=125 280 125 250]{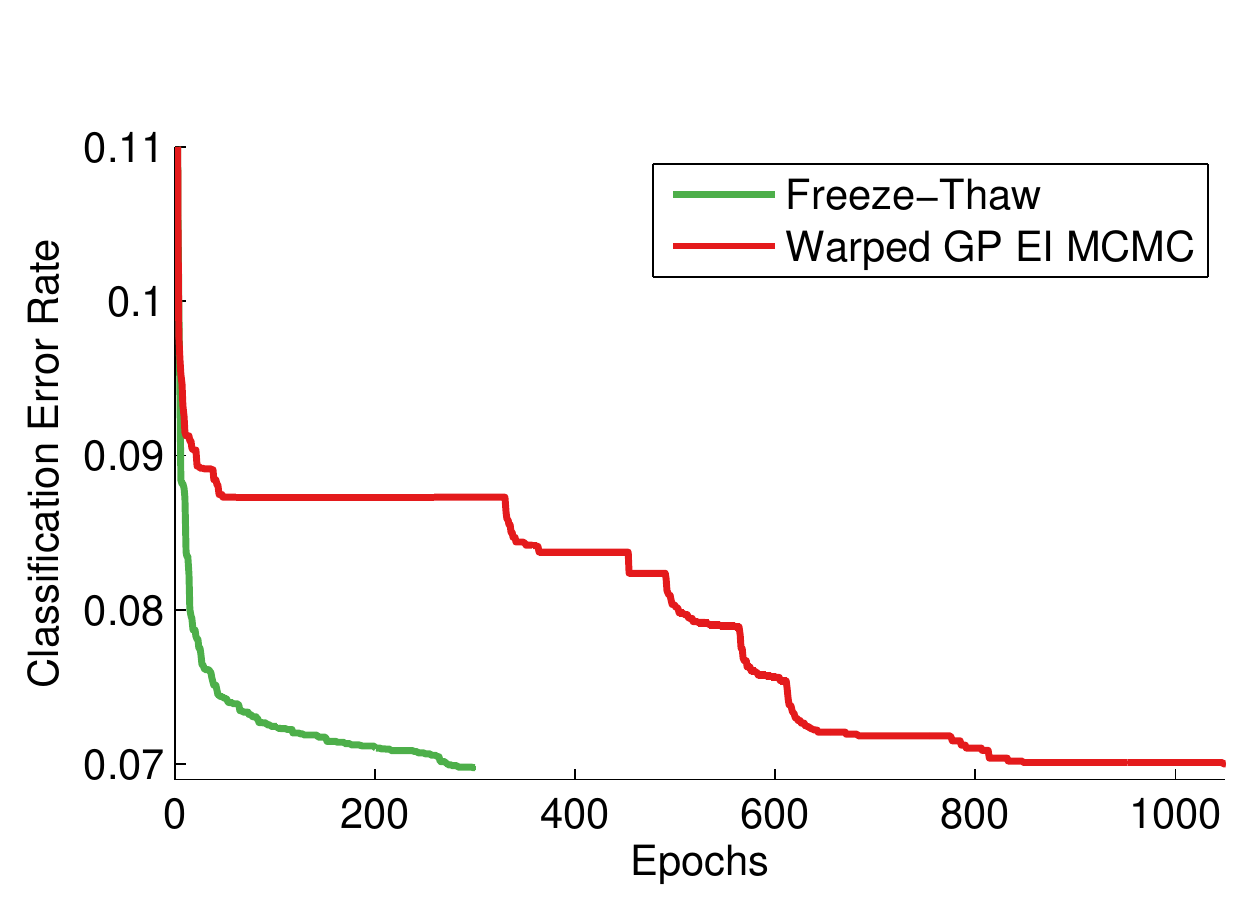}}
\subfloat[\label{fig:lda}Online LDA]{%
  \includegraphics[width=0.32\textwidth,clip=true,trim=125 280 125 250]{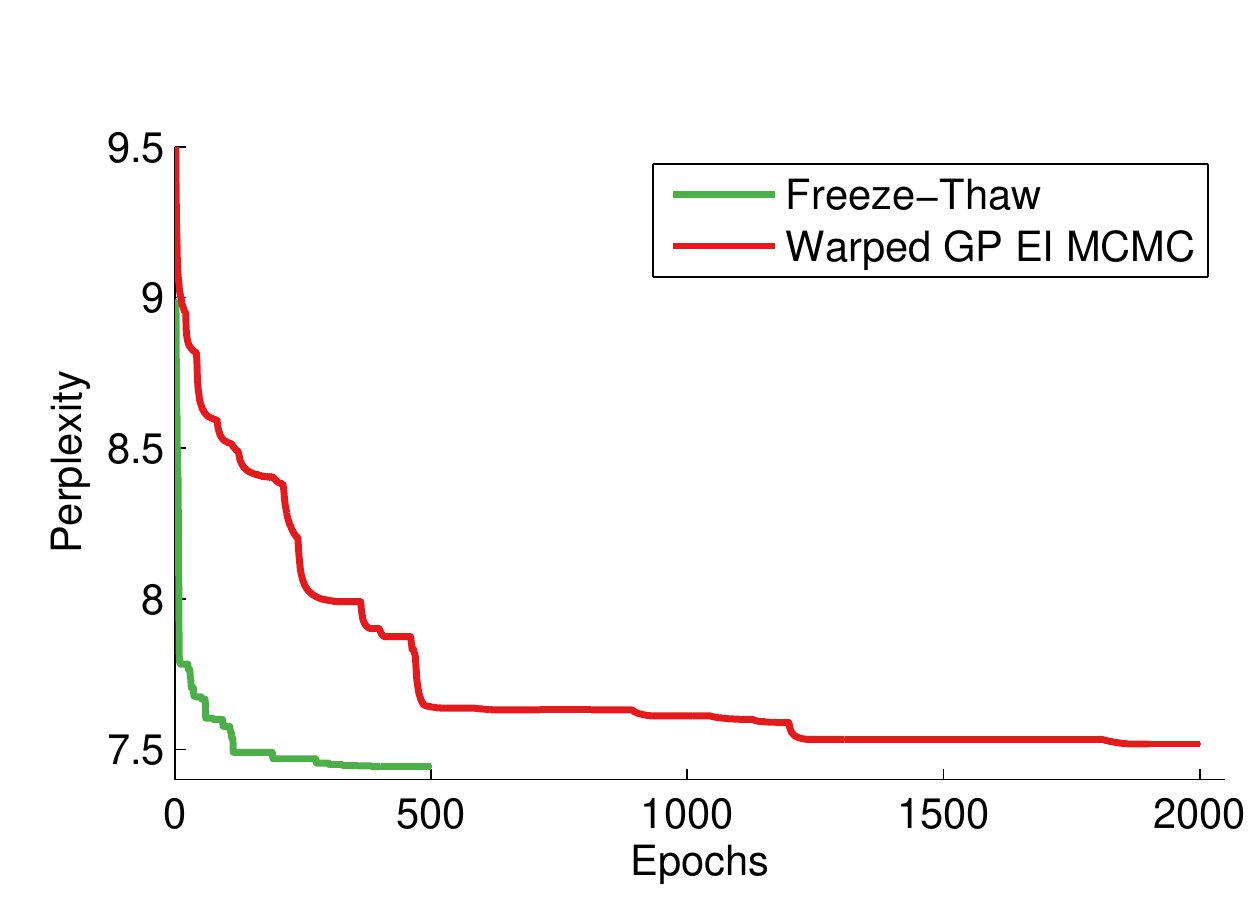}}
\subfloat[\label{fig:pmf}PMF]{%
  \includegraphics[width=0.32\textwidth,clip=true,trim=125 280 125 250]{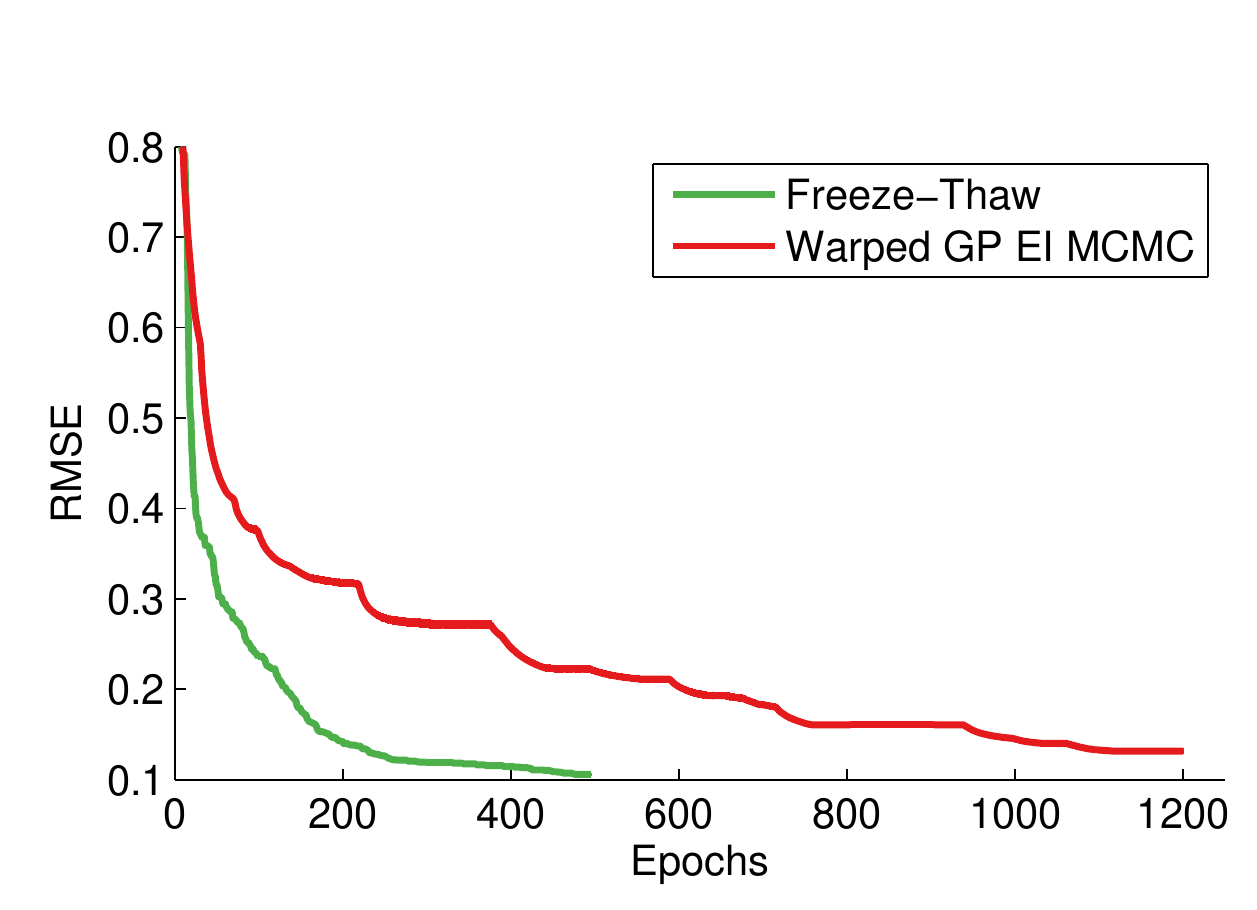}}
\caption{\label{fig:experimental_comparison}This figure shows the results of the empirical comparison to standard Bayesian optimization on three common machine learning hyperparameter optimization problems.  For each problem we report the lowest loss observed over all training epochs evaluated by each method, averaged over five optimization runs.}
\end{figure}

\section{Empirical Analysis}
In this section, we empirically validate our method by comparing to the state-of-the-art (Warped GP EI MCMC) Bayesian optimization method of~\cite{snoek-etal-2014a} on three common machine learning tasks: Online LDA, Logistic Regression and Probabilistic Matrix Factorization.  For each of these tasks, we allowed the method of ~\cite{snoek-etal-2014a} to select the number of training epochs to run, as a hyperparameter to be optimized between 1 and 100, and report at each epoch the cumulative number of epochs run and the lowest objective value observed over all epochs.  We report our results of the comparison in Figure~\ref{fig:experimental_comparison}, visualizing the problem specific loss as a function of the total number of training epochs run throughout each of the Bayesian optimization procedures.  Each experiment was run five times and we report the mean loss.  Both methods used input warping to model non-stationarity. Specific details of our implementation are provided in the appendix.

\paragraph{Logistic Regression}
In the first experiment, we optimize five hyperparameters of a logistic regression trained using stochastic gradient descent on the popular MNIST data set.  The hyperparameters include a norm constraint on the weights (from 0.1 to 20), an $\ell_2$ regularization penalty (from 0 to 1), the training minibatch size (from 20 to 2000), dropout regularization~\cite{hinton2012improving} on the training data (from 0 to 0.75) and the learning rate (from $10^{-6}$ to $10^{-1}$).

\paragraph{Online LDA}
Next we optimize five hyperparameters of an online Latent Dirichlet Allocation (LDA)~\cite{Hoffman2010} experiment on 250,000 documents from Wikipedia.  We optimize the number of topics (from 2 to 100), two Dirichlet distribution prior base measures (from 0 to 2), and two learning rate parameters (rate from $0.1$ to $1$, decay from $10^{-5}$ to 1).  We used the implementation from~\cite{agarwal-etal-2011a} and report average perplexity on a held out validation set of 10\% of the data.

\paragraph{Probabilistic Matrix Factorization}
As a final experiment, we optimize three hyperparameters of a probabilistic matrix factorization (PMF)~\cite{salakhutdinov2008probabilistic} on 100,000 ratings from the MovieLens data set~\cite{herlocker1999algorithmic}.  The hyperparameters include the rank (from 0 to 50), the learning rate (from $10^{-4}$ to $10^{-1}$) and an $\ell_2$ regularization penalty (from 0 to 1).

\begin{figure}[t!]
\centering%
\subfloat[\label{fig:error_curves_pmf_3d}]{%
  \includegraphics[width=0.49\textwidth]{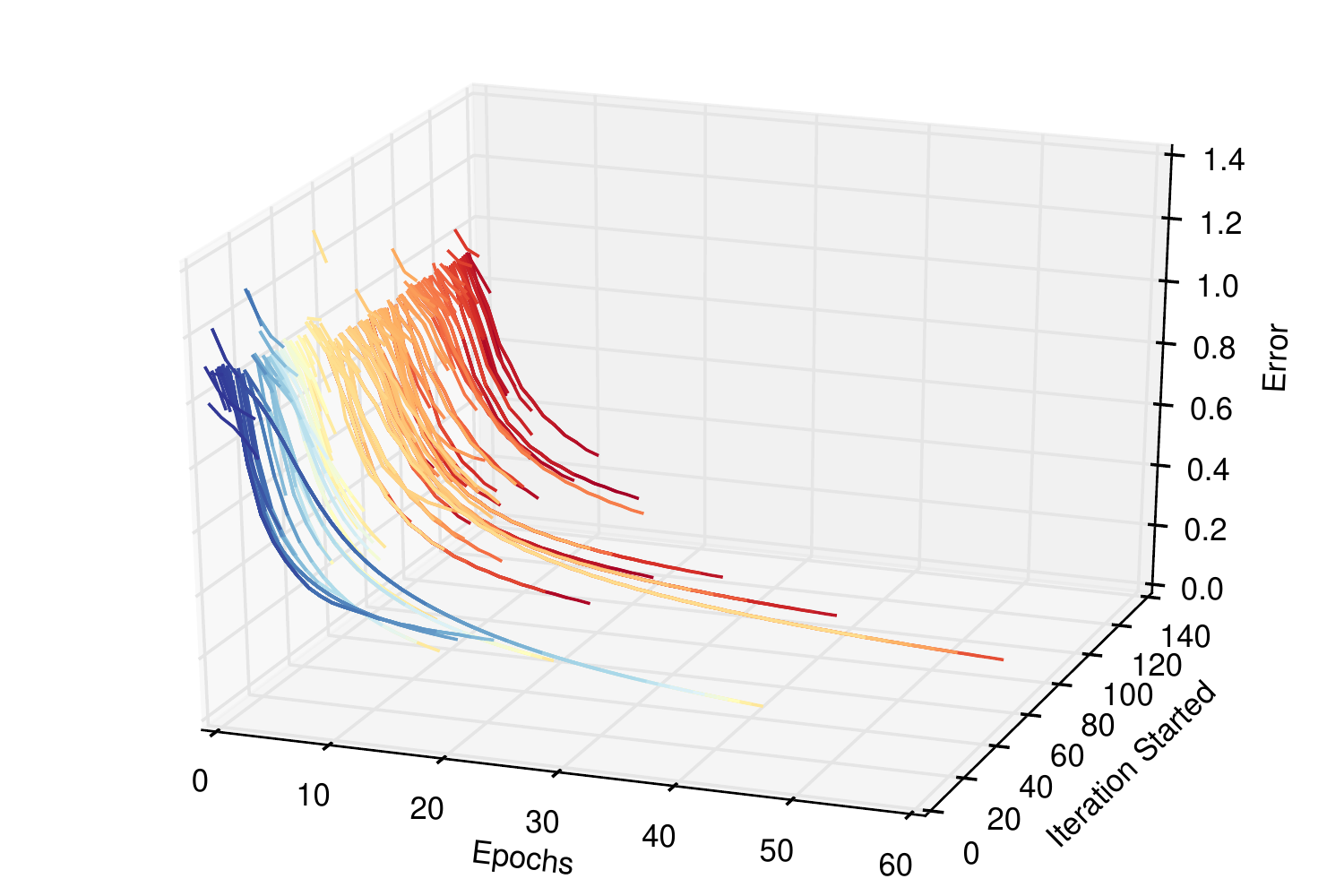}}
\subfloat[\label{fig:error_curves_pmf}]{%
  \includegraphics[width=0.49\textwidth]{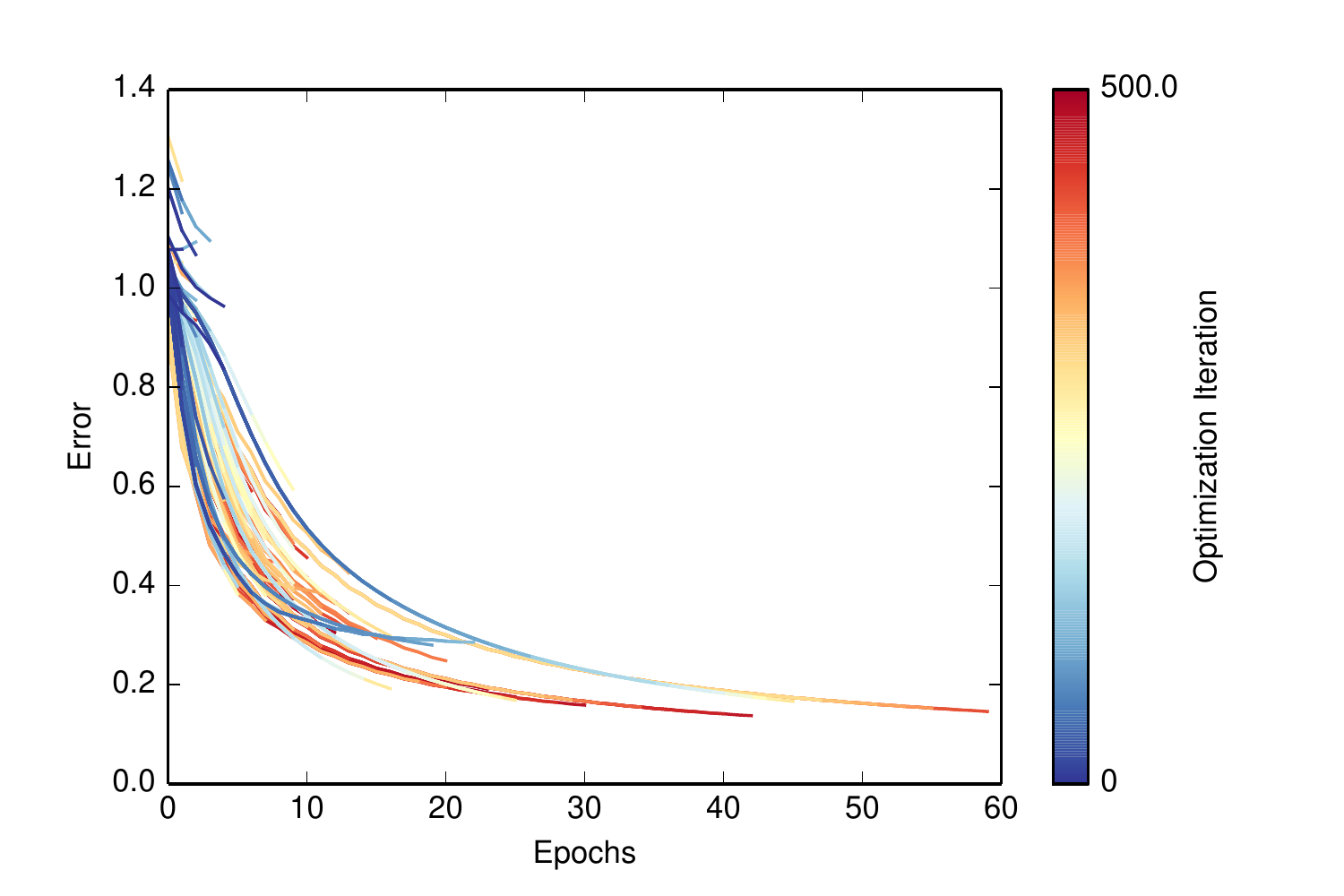}}
\caption{\label{fig:pmf_error_curves}A visualisation of the progression of the optimization curves throughout the optimization procedure on the probabilistic matrix factorization problem.  Figure~\ref{fig:error_curves_pmf_3d} shows for each distinct hyperparameter setting evaluated, the optimization curve run out by the procedure.  The curves are ordered by the iteration in the Bayesian optimization procedure that they were \emph{started} and each epoch of each curve is colored by the iteration of the Bayesian optimization that this section was evaluated. From the figure we can see that the procedure frequently stops running training curves that are not promising and often returns to promising training curves that were previously started.  Figure \ref{fig:error_curves_pmf} shows a two dimensional cross section of~Figure~\ref{fig:error_curves_pmf_3d}.}
\end{figure}

\paragraph{Results}
Figure~\ref{fig:experimental_comparison} shows the results of the empirical analysis in terms of the problem specific loss as a function of the total number of training epochs run out by each method.  Clearly in each of the experiments, our method significantly outperforms the state-of-the-art due to the advantage of being able to dynamically stop and restart experiments.  The difference is particularly prominent for the online LDA problem, where we hypothesize that it is relatively easy to predict the shape of the optimization curves given only a small number of observed epochs.  We assume that the underlying models being optimized are sufficiently expensive that the cost of fitting the GP is negligible.  In practice, the small additional computational effort incurred by our method for explicitly modeling epochs was eclipsed by the performance gains of more rapidly reaching a better loss.

In Figure~\ref{fig:pmf_error_curves}, we show a visualization of the progression of our Bayesian optimization procedure on the PMF problem.  We observe here and throughout the empirical analysis that the method generally initially explored the hyperparameter space by running only a small number of epochs for various hyperparameter settings.  However, once it found a promising curve, it would run it out for more epochs.  Later in the optimization, the method would frequently revisit existing curves and extend them for a few epochs at a time, as we observe in Figure~\ref{fig:error_curves_pmf_3d}.

\section{Conclusion}
Hyperparameter tuning is an important and ubiquitous problem in machine learning that can drastically affect the performance of a model. Given a setting of the hyperparameters, fitting a machine learning model usually involves an iterative training procedure. In this paper, we designed a Bayesian optimization algorithm that is able to exploit the partial information obtained as training proceeds.

Our algorithm relies on a key assumption that training curves tend to follow an exponential decay. We developed a novel, non-stationary kernel as an infinite mixture of exponentially decaying basis functions, and combined this with an efficient temporal Gaussian process prior in order to accurately forecast training curves for a variety of machine learning models. Using an information-theoretic decision framework, our algorithm can dynamically pause, resume, or create new training runs in order to rapidly find good hyperparameter settings.

The methodology developed in this paper can potentially be extended to other problems where partial observations reduce uncertainty. In cases where the exponential decay assumption does not hold, it would be interesting to experiment with other, more flexible priors such as a spatio-temporal GP with a separable covariance.

\section*{Acknowledgements}
This work was funded by DARPA Young Faculty Award N66001-12-1-4219 and an Amazon AWS in Research grant. Jasper Snoek is a fellow in the Harvard Center for Research on Computation and Society.

\appendix
\section*{Appendix}
\section{Efficient Gaussian Processes for Iterative Training Procedures}
\subsection{Inference Derivations}
\paragraph{Posterior distribution}
Using the conditioning property of the Gaussian, we can express the posterior $P(\brmf \given \brmy, \{ \brmx_n \}_{n=1}^N)$ as,
\begin{align}
P(\brmf \given \brmy, \{ \brmx_n \}_{n=1}^N) &= \distNorm \left ( \brmf ; \bmu, \brmC \right ), \nonumber \\
\bmu &= \brmm + \brmK_\rmx \bgamma - \brmK_\rmx \bLambda (\brmK_\rmx^{-1} + \bLambda)^{-1} \bgamma, \nonumber \\
&= \brmm + (\brmK_\rmx^{-1} + \bLambda)^{-1} \bgamma, \nonumber \\
&= \brmm + \brmC \bgamma, \\
\brmC &= \brmK_\rmx - \brmK_\rmx \bLambda \brmK_\rmx + \brmK_\rmx \bLambda (\brmK_\rmx^{-1} + \bLambda)^{-1} \bLambda \brmK_\rmx, \nonumber \\
&= \brmK_\rmx - \brmK_\rmx(\bLambda - \bLambda (\brmK_\rmx^{-1} + \bLambda)^{-1} \bLambda)\brmK_\rmx, \nonumber \\
&= \brmK_\rmx - \brmK_\rmx(\brmK_\rmx + \bLambda^{-1})^{-1} \brmK_\rmx, \\
&= (\brmK_\rmx^{-1} + \bLambda)^{-1}. \nonumber
\end{align}

\paragraph{Posterior predictive distributions}
Given a new hyperparameter setting $\brmx_*$, the posterior distribution $P(f_* \given \brmy, \{ \brmx_n \}_{n=1}^N, \brmx_*)$ is given by,
\begin{align}
P(f_* \given \brmy, \{ \brmx_n \}_{n=1}^N, \brmx_*) &= \int P(f_* \given \brmf, \brmx_*)P(\brmf \given \brmy, \{ \brmx_n \}_{n=1}^N) d\brmf, \nonumber \\
&= \int \distNorm (f_* ; m + \brmK_{\rmx *}^\top \brmK_\rmx^{-1} \brmf, \brmK_{\rmx **} - \brmK_{\rmx *}^\top \brmK_\rmx^{-1} \brmK_{\rmx *}) \distNorm (\brmf ; \bmu, \brmC) d\brmf, \nonumber \\
&= \distNorm (f_* ; m + \brmK_{\rmx *}^\top \brmK_\rmx^{-1} (\bmu - \brmm), \brmK_{\rmx **} - \brmK_{\rmx *}^\top \brmK_\rmx^{-1} \brmK_{\rmx *} + \brmK_{\rmx *}^\top \brmK_\rmx^{-1} \brmC \brmK_\rmx^{-1} \brmK_{\rmx *}), \nonumber \\
&=  \distNorm (f_* ; m + \brmK_{\rmx *}^\top \brmK_\rmx^{-1} (\bmu - \brmm), \brmK_{\rmx **} - \brmK_{\rmx *}^\top (\brmK_\rmx^{-1} - \brmK_\rmx^{-1} \brmC \brmK_\rmx^{-1})\brmK_{\rmx *}), \nonumber \\
&=  \distNorm (f_* ; m + \brmK_{\rmx *}^\top \brmK_\rmx^{-1} (\bmu - \brmm), \brmK_{\rmx **} - \brmK_{\rmx *}^\top (\bLambda - \bLambda \brmC \bLambda)\brmK_{\rmx *}), \nonumber \\
&= \distNorm (f_* ; m + \brmK_{\rmx *}^\top \brmK_\rmx^{-1} (\bmu - \brmm), \brmK_{\rmx **} - \brmK_{\rmx *}^\top (\brmK_\rmx + \bLambda^{-1})^{-1}\brmK_{\rmx *}). \label{eq:postpred}
\end{align}
Similarly, the posterior predictive distribution for a new point in a training curve, $y_{n*}$,  is given by,
\begin{align}
P(y_{n*} \given  \{ \brmx_n \}_{n=1}^N, \brmy) &= \int P(y_{n*} \given \brmy_n, \brmf_n) P(\brmf_n \given \brmy,  \{ \brmx_n \}_{n=1}^N) d\brmf_n, \nonumber \\
&= \int \distNorm(y_{n*} ; \brmf_n \ovec_* + \brmK_{\rmt n *}^\top \brmK_{\rmt n}^{-1} (\brmy_n - \ovec_n\brmf_n), \brmK_{\rmt n **} - \brmK_{\rmt n *}^\top \brmK_{\rmt n}^{-1} \brmK_{\rmt n *}) \nonumber \\ &\qquad\ \distNorm(\brmf_n; \bmu_n, \brmC_{nn}) d\brmf_n, \nonumber \\
&= \distNorm(y_{n*} ; \brmK_{\rmt n *}^\top \brmK_{\rmt n}^{-1} \brmy_n + \Omega \bmu_n, \brmK_{\rmt n **} - \brmK_{\rmt n *}^\top \brmK_{\rmt n}^{-1} \brmK_{\rmt n *} + \Omega \brmC_{nn} \Omega^\top), \label{eq:obsvpred} \\
\Omega &= \ovec_* - \brmK_{\rmt n *}^\top \brmK_{\rmt n}^{-1} \ovec_n. \nonumber
\end{align}
Where $\ovec_*$ is a vector of ones with size equal to the number of time-steps we are predicting. We have omitted the dependence on $t$ for brevity since it is a regularly spaced grid in our experiments.

Finally, in the absence of any observations, the posterior predictive distribution is given by,
\begin{align}
P(y_* \given  \{ \brmx_n \}_{n=1}^N, \brmy, \brmx_*) &=  \int P(y_* \given  f_*) P(f_* \given \{ \brmx_n \}_{n=1}^N, \brmy, \brmx_*)df_*, \nonumber \\
&= \int \distNorm(y; f_*, \brmK_{\rmt**}) \distNorm(f_*; \mu_*, \Sigma_{**})df_*, \nonumber \\
&= \distNorm(y_*; \bmu_*, \brmK_{\rmt **} + \Sigma_{**}).
\label{eq:noobsvpred}
\end{align}
Where $\mu_*$ and $\Sigma_{**}$ are the mean and variance given by Equation \ref{eq:postpred}.
\section{Implementation Details}
In our experiments we follow the conventions laid out in~\cite{snoek-etal-2012b} and use a modification of it's accompanying Spearmint package~\footnote{\url{https://github.com/JasperSnoek/spearmint}}. The specific details of our GP model implementation are given below.
\subsection{Kernels and GP Hyperparameters}
We use a Mat\'{e}rn-\sfrac{5}{2} to determine the function over hyperparameters, along with the warping technique developed in~\cite{snoek-etal-2014a}
\begin{align*}
k_{\mathrm{M}52}(w(\brmx),w(\brmx')) &= \theta_0 \left (1 + \sqrt{5r^2} + \frac{5}{3}r^2\right )\exp \left( -\sqrt{5r^2} \right), \\
r^2 &= \sum_{d=1}^{D}\frac{(w_d(x)_d - w_d(x')_d)^2}{\theta_d^2}, \\
w_d(x) &= \mathrm{BetaCDF}(x,a_d,b_d).
\end{align*}
Where BetaCDF refers to the cumulative distribution of the beta distribution with shape parameters $a$ and $b$.
We place a log-normal prior with a log-scale of $0$ on $\theta_0$, $a_d$, and $b_d$ for all $d$, and a uniform prior on $\theta_d$,
\begin{align*}
\theta_0 &\sim \mathrm{lognorm}(0,1), \\
a_d &\sim \mathrm{lognorm}(0,1)\quad d=1...D, \\
b_d &\sim \mathrm{lognorm}(0,1)\quad d=1...D, \\
\theta_d &\sim \mathrm{uniform}(0,10).
\end{align*}
For the kernel over epochs, we use our custom exponential decay kernel along with an additive noise kernel.
\begin{align*}
k_{\text{exp decay}}(t,t') &= \frac{\beta^\alpha}{(t + t' + \beta)^\alpha} + \delta(t,t')\sigma^2.
\end{align*}
We place a lognormal prior on the hyperparameters $\alpha$ and $\beta$, and a horseshoe prior on $\sigma^2$.
\begin{align*}
\alpha &\sim \mathrm{lognorm}(0,1)\quad d=1...D, \\
\beta &\sim \mathrm{lognorm}(0,1)\quad d=1...D, \\
\sigma^2 &\sim \mathrm{horseshoe}(0.1)~\text{\cite{carvalho-2009a}}.
\end{align*}

Finally, we use a constant prior mean $m$ with a uniform hyperprior for the GP over hyperparameters. We ensure that this value does not exceed the bounds of the observations.
\begin{align*}
m &\sim \mathrm{uniform}(y_\mathrm{min},y_\mathrm{max}).
\end{align*}

\bibliographystyle{plain}
\bibliography{freeze-thaw-nips2014}
\normalsize

\end{document}